\documentclass[conference]{IEEEtran}
\IEEEoverridecommandlockouts
\usepackage{cite}
\usepackage{amsmath,amssymb,amsfonts}
\usepackage{algorithmic}
\usepackage{graphicx}
\usepackage{textcomp}
\usepackage{xcolor}
\graphicspath{{images/}}
\usepackage{multirow}
\usepackage{subfig}
\usepackage{multicol}

\def\BibTeX{{\rm B\kern-.05em{\sc i\kern-.025em b}\kern-.08em
    T\kern-.1667em\lower.7ex\hbox{E}\kern-.125emX}}
\begin{document}

\title{Enhancing Fine-Grained Classification for Low Resolution Images}

\author{\IEEEauthorblockN{Maneet Singh$^1$, Shruti Nagpal$^1$, Mayank Vatsa$^2$, and Richa Singh$^2$}
\IEEEauthorblockA{$^1$IIIT-Delhi, India; $^2$IIT Jodhpur, India \\
Email: \{maneets, shrutin\}@iiitd.ac.in, \{mvatsa, richa\}@iitj.ac.in}
}

\maketitle

\begin{abstract}
Low resolution fine-grained classification has widespread applicability for applications where data is captured at a distance such as surveillance and mobile photography. While fine-grained classification with high resolution images has received significant attention, limited attention has been given to low resolution images. These images suffer from the inherent challenge of limited information content and the absence of fine details useful for sub-category classification. This results in low inter-class variations across samples of visually similar classes. In order to address these challenges, this research proposes a novel attribute-assisted loss, which utilizes ancillary information to learn discriminative features for classification. The proposed loss function enables a model to learn class-specific discriminative features, while incorporating attribute-level separability. Evaluation is performed on multiple datasets with different models, for four resolutions varying from $32\times 32$ to $224\times 224$. Different experiments demonstrate the efficacy of the proposed attribute-assisted loss for low resolution fine-grained classification.

\end{abstract}


\section{Introduction}

%Fine-grained classification focuses on differentiating between objects belonging to distinct categories of the same parent class. That is, it operates at the sub-category level for a particular class. For example, distinguishing between different types of birds \cite{cub}, animals \cite{awa2}, cars \cite{cars}, or clothes \cite{deepFashion}. Fine-grained classification is useful in scenarios of similar sample retrieval, such as online shopping or recommendation systems, automated vertebrate identification (in case of birds or animals sightings), and vehicle make/model recognition for automated toll systems. Owing to the complex nature of the problem, fine-grained classification has garnered significant attention over the past few years. This has resulted in several challenging real-world datasets and algorithms for addressing the given problem. 

\begin{figure}[!ht]
\centering
\subfloat[High and low resolution samples from the AwA-2 dataset \cite{awa2}]{\includegraphics[width=3.2in]{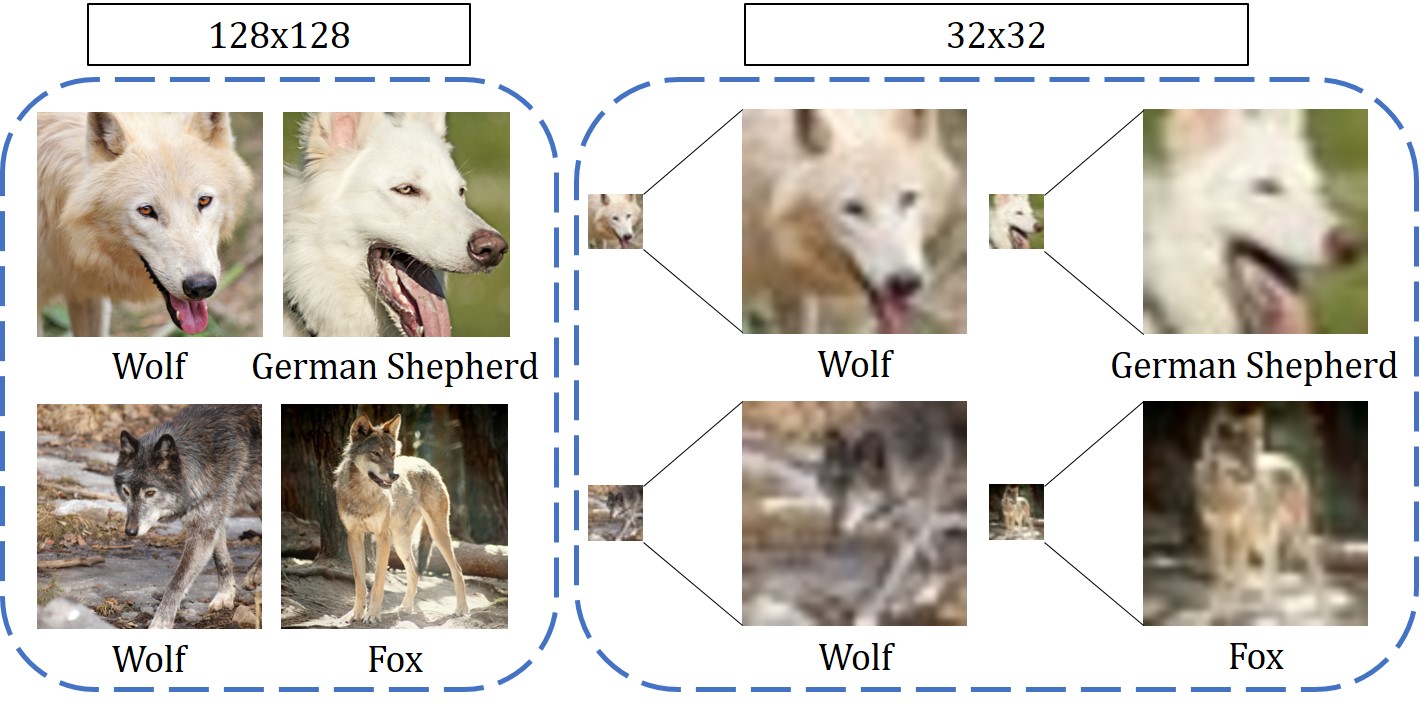}}\\
\subfloat[Learned space via the proposed attribute-assisted loss]{\includegraphics[width=3.2in]{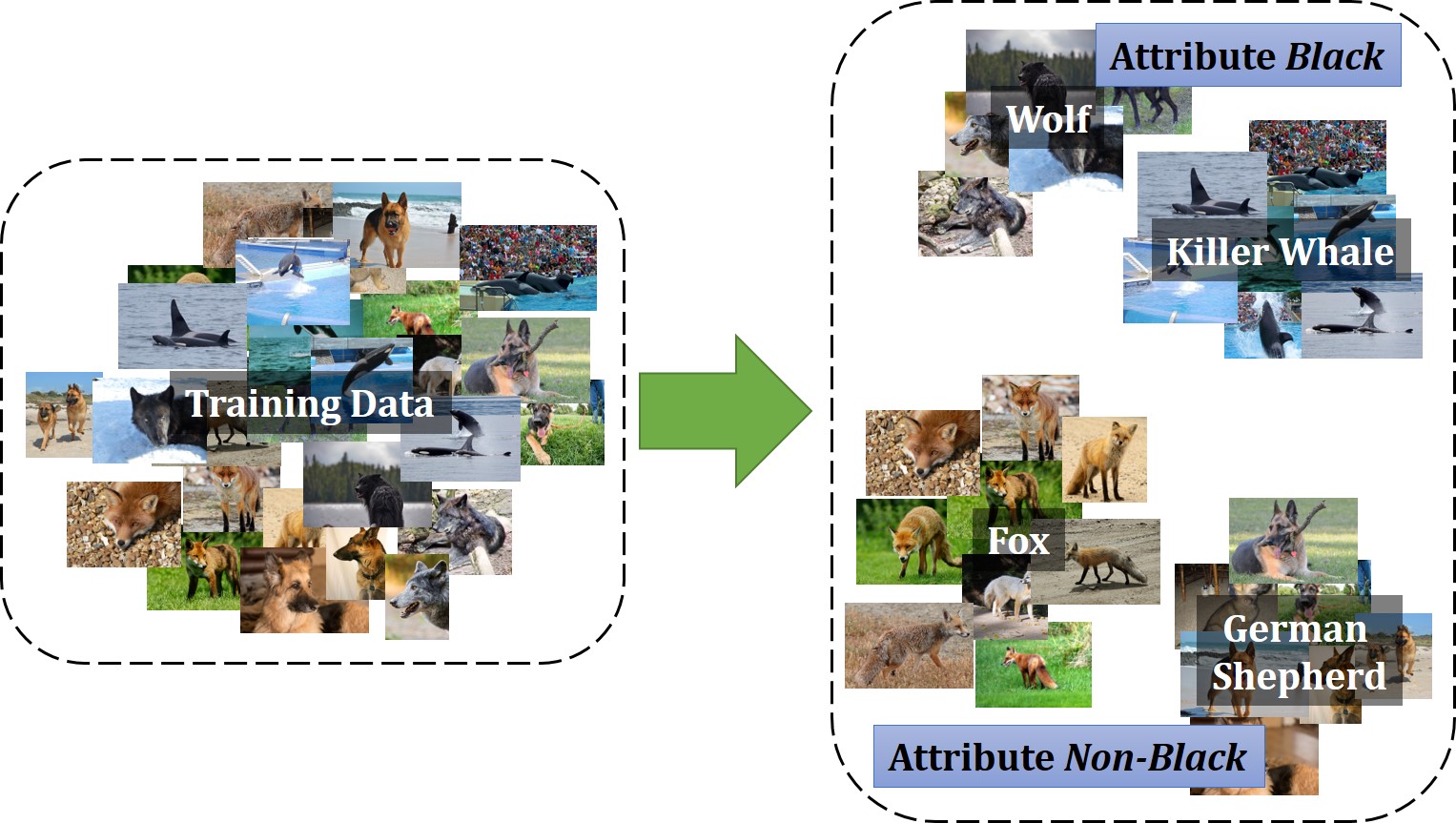}}
\caption{(a) It is difficult to distinguish between data belonging to the same parent class, even at a high resolution; the problem is further aggravated in low resolution samples. (b) The proposed attribute-assisted loss introduces a higher level of separation based on the given attribute (\textit{black}), followed by class-wise separability. Best viewed in color.}
\vspace{-10pt}
\label{fig:intro}
\end{figure}

\begin{figure*}[!ht]
\centering
\includegraphics[width=6.5in]{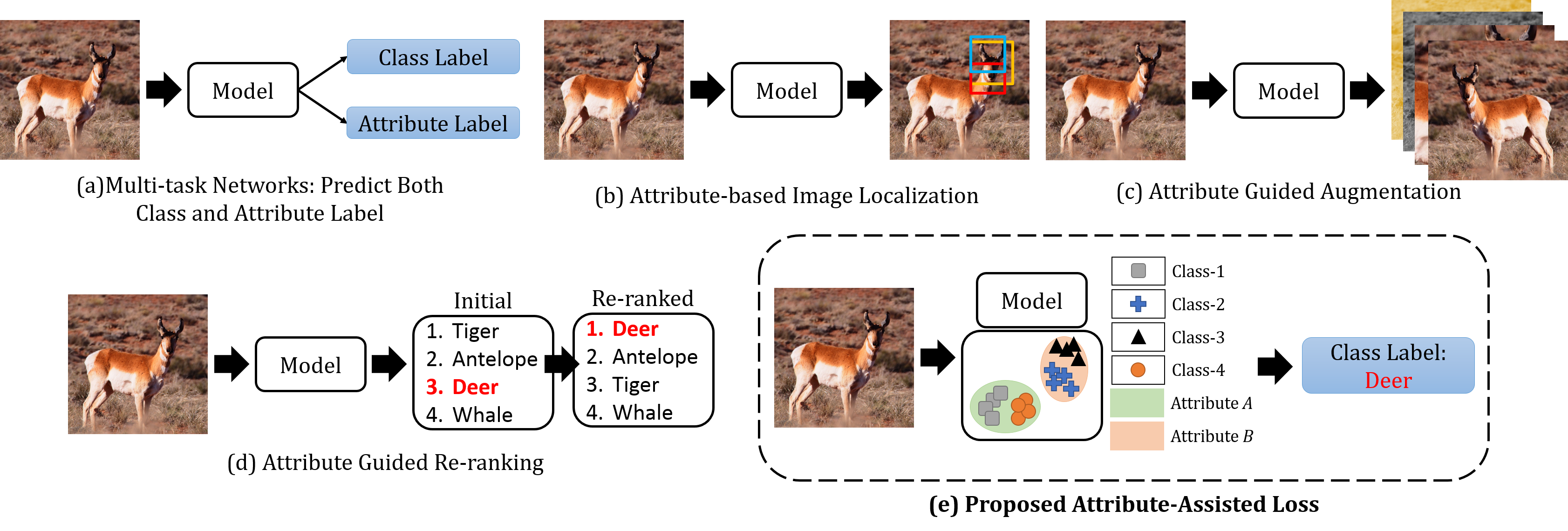}
\caption{(a)-(d): In the literature, researchers have utilized attribute information for fine-grained classification; (e): the proposed attribute-assisted approach. Different from existing techniques, the proposed loss implicitly utilizes the attributes to learn a more discriminative space, as opposed to predicting attributes or explicitly using them for ancillary tasks. Best viewed in color.}
\label{fig:intro1}
\vspace{-10pt}
\end{figure*}

Fine-grained classification focuses on distinguishing between objects belonging to distinct categories of the same parent class. Most of the existing research has focused on high resolution images \cite{awa2,cars,deepFashion,cub,ijcnn2019,ijcnn2020, singhPAMI}, without much attention to low resolution data. Fine-grained classification of low resolution images is useful in surveillance scenarios, where low resolution data is captured from CCTV cameras. If a crime is captured on camera, clothes identification or vehicle make/model recognition is often important for obtaining supplementary information about suspects. The information can be circulated to create awareness or aid automated recognition. Other than surveillance, low resolution classification also finds applicability in data captured via smartphones containing objects of relatively smaller resolution. As observed in the literature, classification models trained on high resolution images often do not perform well on low resolution samples \cite{singh2018identity}, thereby creating a need for algorithms designed specifically for low resolution fine-grained classification \cite{bhattTIP,prl17_cai,icip15}.  

%Despite the wide-spread utility of low resolution fine-grained classification, the problem has received limited attention, with few algorithms designed specifically for it \cite{prl17_cai,icip15}.  

%The challenge of fine-grained classification is further aggravated in low resolution images. 
%Since fine-grained classification focuses on sub-categories, it suffers from the challenge of low inter-class variations, thereby rendering it an arduous task. This problem is further aggravated for low resolution images.Figure \ref{fig:intro} demonstrates the additional challenges observed for the given task. 
%In order to perform classification in such scenarios, the aim is to develop a model which incorporates additional attribute-based information and learns discriminative features for the given task. In such scenarios, the aim is to develop a model which is able to learn discriminative features, 

Fig. \ref{fig:intro}(a) presents sample high and low resolution images from the Animals with Attributes-2 \cite{awa2} dataset. Low resolution samples are shown at $32\times32$ resolution, along with their corresponding bicubic interpolations. It is interesting to note that due to the similarity in structure and features, even at the high resolution of $128\times128$, it can be difficult to distinguish between the two images present in a row (such as Wolf and German Shepherd, or Wolf and Fox). In such cases, one generally utilizes fine details of localized regions such as the nose shape or eye color to supplement classification. However, at lower resolutions such as $32\times32$, these fine details are often lost, making the images almost indistinguishable. Therefore, in such scenarios, additional information which may or may not be present visually, can be utilized to aid recognition and enhance classification performance.
Inspired by these observations, as shown in Fig. \ref{fig:intro}(b), this paper presents a novel \textit{attribute-assisted loss} for low-resolution fine-grained classification. The proposed technique is built upon the philosophy of preventing attribute-level mis-classifications. For instance, if a low resolution image of a female subject is given as query, the output should be similar looking faces. In other words, a trained system should be able to differentiate with respect to the inherent variations observed due to the gender attribute and return faces belonging to the `female' attribute. %Since attributes can provide a higher level separation between visually similar data, we propose incorporating the information for enhancing fine-grained classification. %For the example of Wolf and Fox demonstrated in Figure \ref{fig:intro}, the model should discriminate between animals which are \textit{black} and \textit{non-black} (Figure \ref{fig:intro} (b)). 
Similarly, for the case of Wolf and German Shepherd, the model should clearly distinguish between \textit{forest} and \textit{non-forest} animals at a conceptual level\footnote{These attributes are provided in the Animals with Attributes-2 dataset.}. Since attributes can provide a higher level separation between visually similar data, we propose incorporating them for enhancing fine-grained classification. Different from the existing techniques which explicitly utilize attributes for fine-grained classification, as shown in Fig. \ref{fig:intro1}(a-d), the proposed attribute-assisted loss implicitly uses the attribute information for learning a better classification space.
%The proposed attribute-assisted loss enforces these constraints by combining an attribute-oriented term with a classification loss. 
%That is, the proposed loss assists the network for classification by incorporating the discrimination observed on the attribute level. 
The contributions of this research are:

\begin{itemize}
\item We present a novel attribute-assisted loss for low resolution fine-grained classification. The proposed loss is model agnostic and utilizes attributes as ancillary information such that a network learns category-level discriminative features while incorporating attribute based similarity. To the best of our knowledge, this is the first research which utilizes attributes as ancillary information (which may or may not be visually present in the image) for aiding low resolution fine-grained classification. 
\item The effect of learning a network with both visual and non-visual (conceptual) attributes has also been studied in this research. Since the network learns an attribute-assisted discriminative feature space, improved performance is observed across both kinds of attributes.
\item To the best of our knowledge, this is amongst the first few works which performs an in-depth evaluation for low resolution fine-grained classification, spanning across multiple resolutions of $32\times32$, $48\times48$, and $64\times64$, with three state-of-the-art deep learning models. Experiments on two publicly available datasets demonstrate the efficacy of the proposed loss.
\end{itemize}

\begin{figure*}
\centering
\includegraphics[width=6.8in]{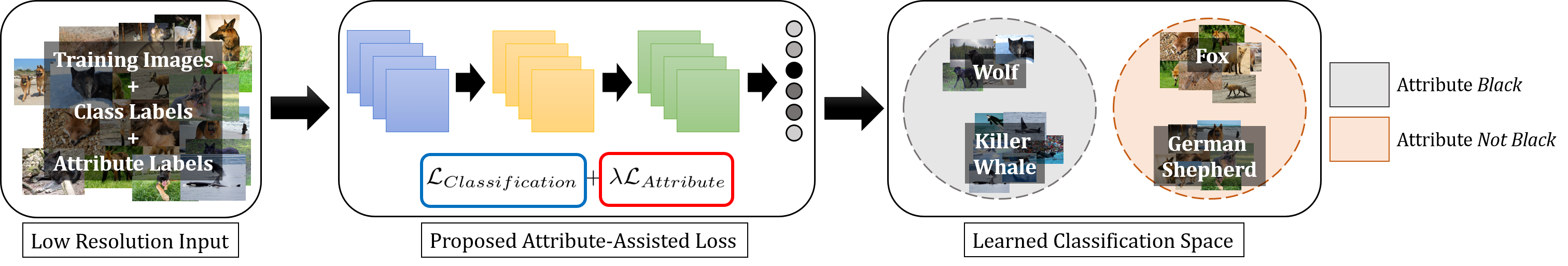}
\caption{Proposed attribute-assisted loss facilitates learning of a discriminative space for fine-grained classification. A higher level separability is enforced between classes having different attributes (eg. \textit{black} and \textit{not black}), while the trained model is able to distinguish between classes of the same attribute as well (eg. \textit{Fox} and \textit{German Shephard}). Images have been taken from the AwA-2 dataset \cite{awa2}.}
\label{fig:algo}
\vspace{-10pt}
\end{figure*}

\section{Related Work}

Majority of the research in fine-grained classification has focused on high resolution images \cite{survey17}, with tremendous focus on part detection, visual attention, and localization techniques \cite{fu17Cvpr,he17AAAI,oh2020salient,liu17AAAI,peng18Tip,hanselmann2020elope,yang2018learning,zhao17TMM,zheng17Iccv}. Such methods isolate regions of the input image which contain distinctive characteristics useful for classification. For example, in case of bird classification, models are often trained to locate the wings, beak, or tail, in order to identify distinguishing features. Apart from attention or localization based networks, researchers have also proposed modifications to the Convolutional Neural Network architecture such as the Channel Max Pooling \cite{ma2019fine}, Grassmann Pooling \cite{wei2018grassmann}, Low-rank Bilinear Pooling \cite{kong17Cvpr}, or Alpha Pooling \cite{simon17Iccv}, specifically for the task of fine-grained classification. Techniques which utilize additional information or encode features in a hierarchical manner have also shown to perform well \cite{cai17Iccv,gebru17Iccv}. 

\noindent\textbf{Low Resolution Fine-Grained Classification:} The problem of low resolution fine-grained classification has received limited attention. In 2015, Chevalier \textit{et al.} \cite{icip15} studied the effect of decreasing the resolution of images for the task of fine-grained classification. They evaluated the performance of Fisher vectors and deep learning based representations for the given task. It was observed that the performance of fine-grained classification degrades rapidly below the resolution of $50\times50$. As demonstrated in Fig. \ref{fig:intro}(a), low resolution images often lack fine details, thereby rendering the task of part detection or localization challenging. In order to overcome this challenge, Cai \textit{et al.} \cite{prl17_cai} proposed an end-to-end Convolutional Neural Network (CNN) based model for classification. The authors demonstrate improved performance in comparison with other techniques, by introducing an additional step of super resolution before classification. Often for smaller resolutions, the task of super-resolution is deemed challenging, since it requires synthesizing images of higher dimensions, with a large magnification factor, from limited information content.

\noindent\textbf{Attributes for Fine-Grained Classification:} As shown in Fig. \ref{fig:intro1}, attribute information has widely been used for fine-grained classification for high resolution images. The existing techniques can broadly be categorized as follows: (a) multi-task networks \cite{iccv15,zhangAttr}; (b) algorithms which utilize attribute-based image localization, i.e. training networks \textit{where} to look in an image \cite{gebru17Iccv,liu17AAAI,yan17}; (c) attribute-guided augmentation; or (d) attribute guided re-ranking of outputs \cite{attrAug2,attrAug1}. 
%In literature, attributes have been used to perform part localization in high resolution images, i.e. training networks \textit{where} to look in an image \cite{gebru17Iccv,liu17AAAI,yan17}. Attributes have also been used for augmentation by increasing the class-specific training images, or performing re-ranking at the classification level \cite{attrAug2,attrAug1}. Often, attribute prediction is added as a component in the classification network, resulting in a multi-task framework \cite{iccv15,zhangAttr}. 
Most of these techniques either explicitly incorporate the visual attribute information in the classification network, or utilize an attribute prediction component. While this is feasible for high resolution images with high information content, we believe it is challenging for low resolution data. Therefore, in this work, we propose to utilize attributes as additional information, capable of assisting the network for discriminative features.

%\begin{figure}
%\centering
%\subfloat[][Training Stage]{\includegraphics[width = 3.2in]{train.jpg}} \\
%\subfloat[][Testing Stage]{\includegraphics[width = 3.2in]{test.jpg}} 
%\caption{Proposed attribute-assisted loss.}
%\label{fig:algo}
%\end{figure}

\section{Proposed Attribute-Assisted Loss}

%Most of the state-of-the-art algorithms for fine-grained classification are designed explicitly for high resolution images, and utilize specific visual cues to enhance the classification performance. This is often achieved via part detection or localization on the input image. Owing to the lack of information content in low resolution images (Fig. \ref{fig:intro}), it is our hypothesis that identifying such visual cues or fine details would be challenging. Therefore, instead of relying on the visual information only, this research presents a novel attribute-assisted loss for fine-grained image classification. The proposed technique introduces soft discrimination at a higher level via attributes, to aid the task of classification. 

\begin{figure*}
\centering
\includegraphics[width=6.4in]{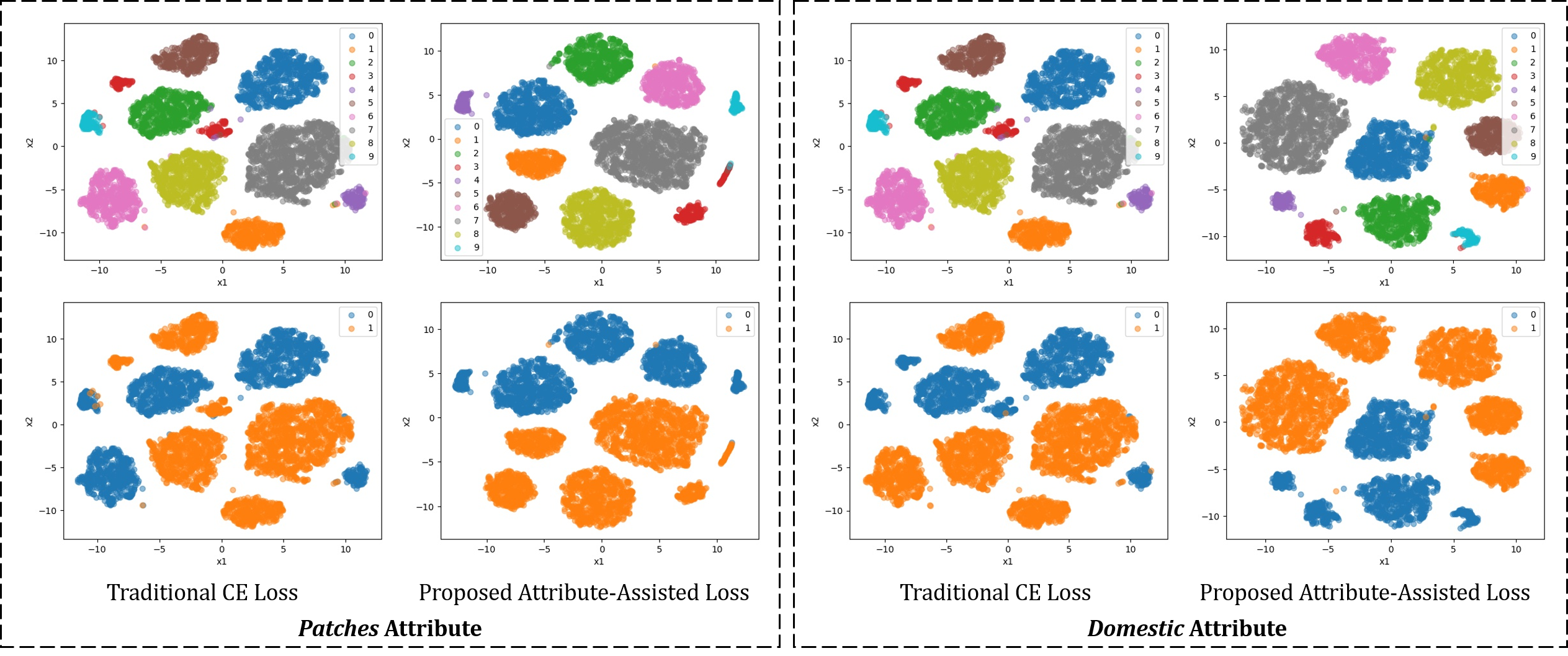}
\caption{t-SNE plot of learned features on a subset of the Animals with Attributes-2 dataset, containing 10 classes. The first row corresponds to the distribution of the learned representation with respect to the class labels, and the second row provides the attribute labels of the data points. It can be observed that attribute-assisted loss forces samples with the same attribute to be together, while the classification term enforces discrimination between the classes. Best viewed in color.}
\label{fig:visualization}
\end{figure*}

Fig. \ref{fig:algo} presents a broad overview of the proposed attribute-assisted loss, applicable to any \textit{representation learning} algorithm. The proposed loss utilizes attribute information at the time of training in order to provide a higher-level of discrimination for low resolution fine-grained classification. Once the model is trained, it is used for classification on the test images, without requiring any attribute information. The proposed technique contains two components: (i) classification loss, and (ii) attribute loss, mathematically expressed as: 
\begin{equation}
\mathcal{L}_{Proposed} = \mathcal{L}_{Classification} + \lambda\mathcal{L}_{Attribute}
\label{genericEq}
\end{equation}
where, $\lambda$ denotes the weight given to the attribute loss. The classification loss is used for learning the discriminative class-specific boundaries, while the attribute loss incorporates a higher level of separability between classes. The proposed attribute-assisted loss utilizes a triplet of training data: $\{x_i, y_i, a_i\}$, where $x_i$ refers to the $i^{th}$ training image, $y_i$ refers to its class label, and $a_i$ refers to the attribute label of the $i^{th}$ sample. For $n$ training samples, the proposed \textit{attribute-assisted loss} is mathematically formulated as:
\begin{equation}
\mathcal{L}_{Proposed} = -\frac{1}{n}\sum_{i=1}^{n}y_i\log(p_{y_i}) \ + \ \lambda \frac{1}{2}\sum_{i=1}^{n} \|f(x_i) - r(a_i) \|_2^2
\label{eq:prop}
\end{equation}
where, $y_i$ is the ground-truth class label for the $i^{th}$ sample, and $p_{y_i}$ refers to the prediction given by the model (a scalar value). $f(x_i)$ refers to the representation learned by the model for $x_i$ (often of length $512$ or $1024$), and $r(a_i)$ refers to the representative feature for the scalar attribute $a_i$. In the above Equation, the first term corresponds to the cross-entropy loss used for classification, while the second term corresponds to the additional proposed attribute loss, which reduces the intra-class variations with respect to the attribute information. While models trained with the cross-entropy loss only are useful for classification, however, in the scenario of low resolution fine-grained classification, they might not be able to learn meaningful representations due to the inherent lack of discriminative information content. It is our assertion that attribute information introduced via an additional attribute loss can help in learning better models, thus improving the classification performance. 

For low resolution classification, where the model works with images having limited information, the additional attribute loss presents a two-fold purpose: (i) attribute labels are provided as additional information during training, thus increasing the available information for learning a discriminative model, (ii) the additional information facilitates learning of attribute encoded features such that the model does not confuse between samples being \textit{conceptually} different; or in case of visual attributes, it forces the network to learn representations capable of discriminating on the given label. Considering the example of animal classification in low resolution images, the recognition framework should not confuse between animals found in \textit{forests} with those found in \textit{water}, and between animals \textit{with} or \textit{without patches}. The attribute loss facilitates feature learning such that the data corresponding to the same attribute label is brought closer, thus reducing the attribute-level mis-classifications. For an input $x_i$, this is achieved by minimizing the distance between the learned representation ($f(x_i)$) and the representative feature of its attribute, calculated as the mean feature of all samples with attribute $a_i$ \cite{centerLoss}. Mathematically, the attribute loss is formulated as: 
\begin{equation}
\label{center}
\mathcal{L}_{Attribute} = \frac{1}{2}\sum_{i=1}^{n} \|f(x_i) - r(a_i) \|_2^2; \ r(a_i) = \mu(f(\mathbf{X}_{a_i})) 
\end{equation}
where, $\mathbf{X}_{a_i}$ refers to all training samples having attribute $a_i$, and $\mu$ refers to the mean operator. Since the attribute loss is in conjunction with a classification loss, this enables the network to learn class-specific discriminative features, while incorporating the attribute-level separability as well. Attribute information is provided as input at the training time, and is thus not predicted from the network. This enables the model to learn attribute-assisted features useful for classification, while eliminating the need for attribute labels during testing.

\begin{figure*}
\centering
\subfloat[][Animals with Attributes-2 Dataset]{\includegraphics[trim = {0in, 0.1in, 0in, 0in}, clip, width = 3.2in]{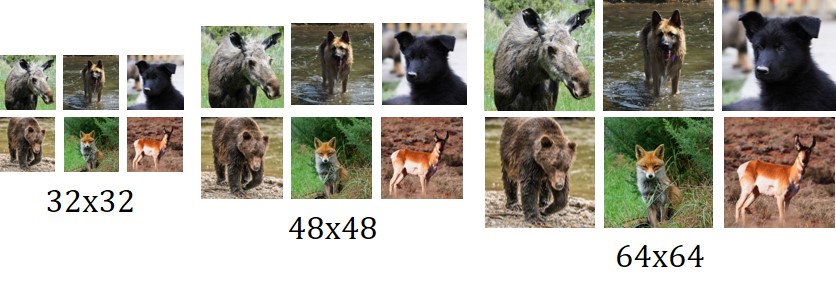}} 
\hspace{0.5em}
\subfloat[][DeepFashion Dataset]{\includegraphics[trim = {0in, 0.1in, 0in, 0in}, clip, width = 3.2in]{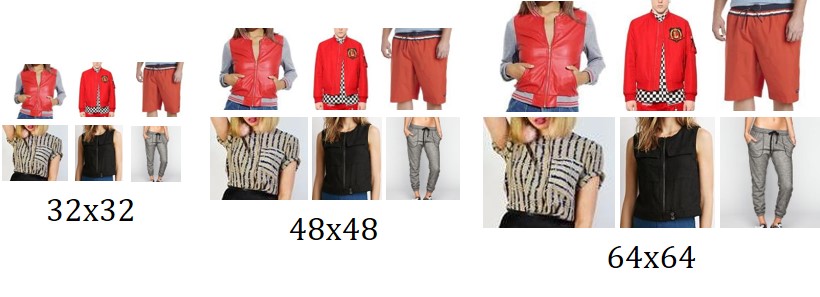}}
\caption{Sample images from the datasets used for experiments, for three different resolutions.}
\label{fig:sample}
\vspace{-5pt}
\end{figure*}

As a toy example, the benefit of learning a network via the attribute-assisted loss has visually been presented in Fig. \ref{fig:visualization}. A subset of Animals with Attributes-2 (AwA-2) dataset \cite{awa2} containing 10 classes is randomly chosen for this experiment, where 70\% of each class is used for training, while the remaining samples form the test set. Pre-trained ResNet-18 \cite{resnet} is trained using the traditional Cross Entropy (CE) loss and the proposed attribute-assisted loss for $32\times32$ resolution images. Visualizations are shown for two attributes: \textit{patches} and \textit{domestic}. The figure shows the final layer representations, projected onto a 2-D space using t-SNE \cite{tsne}. It is interesting to observe that while both the models are able to discriminate well between the 10 classes, the models learned via the attribute-assisted loss are also able to differentiate the features based on the attribute. Classes belonging to a particular attribute are grouped together, while ensuring separability at the class-level. In this example, the attribute-assisted loss demonstrates an improvement of over 5\% on the test set. This shows the benefits of the attribute-assisted loss for low resolution fine-grained classification.  

%\noindent \textbf{Training and Testing:}
As shown in Fig. \ref{fig:algo}, a model trained with the proposed loss utilizes the image samples, class labels, and the attribute labels while training. Model optimization is performed on Eq. \ref{eq:prop} for learning discriminative features, which encode the attribute information as well. During testing, the learned network utilizes only the input image, to predict the class label. Ground truth attribute label is not required for testing, nor does the model predict the attribute of the input sample.

%Here, for the example of Convolutional Neural Network (CNN), a traditional architecture contains a softmax layer as the final layer, which is used for performing classification. 
%In order to complement the learning of discriminative features for the task of low resolution fine-grained classification, the proposed technique contains two components: (i) classification based loss, and (ii) attribute based loss. That is, 

\section{Experiments and Implementation Details}
\label{sec:protocol}
The proposed approach is evaluated on two publicly available datasets: Animals with Attributes-2 \cite{awa2} and DeepFashion \cite{deepFashion} (Fig. \ref{fig:sample}). The details of these datasets are: 

%Both the datasets have been used for performing fine grained classification, details regarding each are as follows:

\noindent \textbf{AwA-2 dataset} \cite{awa2} contains 37,322 images of animals belonging to 50 categories. Due to the lack of an existing protocol for fine-grained classification, we define the training and testing partitions. 70\% data from each class is randomly chosen to form the training set, and the remaining samples form the test set. One sample of each class from the training set is randomly selected to create the validation set. The dataset also contains 85 binary attributes for each class, such as \textit{white}, \textit{domestic}, and \textit{patches}. During training, data augmentation has been performed by random horizontal flipping.

%Images of the AwA-2 dataset have been used as it is for experiments, without any bounding box based cropping.
 
\noindent \textbf{DeepFashion dataset} \cite{deepFashion} contains over 8,00,000 images of clothes captured in different settings with varying pose, illumination, and resolution. A pre-defined protocol is provided for clothing categorization, containing 289,219 images of 46 classes. The testing and validation set contain 40,000 images each, and the training set contains the remaining 209,219 images. Each image has been annotated with 1,000 attributes, along with a bounding box. Each class is also annotated with an additional attribute of \textit{upper-body}, \textit{lower-body}, or \textit{full-body}. Since the proposed technique utilizes class-level attributes, the image-specific information has not been used in the experiments. %The bounding boxes have been used to crop the images in order to remove the additional background information.\\ 

The efficacy of the proposed attribute-assisted loss has been demonstrated by performing experiments on three different models:  ResNet-18, ResNet-50 \cite{resnet}, and DenseNet-121 \cite{densenet}. All three models have shown to achieve superior performance on the ImageNet dataset \cite{imagenet}, which contains over 3.2 million images of different objects. To demonstrate the effectiveness of the proposed loss, results have been computed by three methods: (i) extracting features from the pre-trained network and classifying using Euclidean distance, (ii) fine-tuning the pre-trained network using the standard Cross Entropy loss and performing classification, (iii) fine-tuning the pre-trained network using the proposed attribute-assisted loss and performing classification. The models have been evaluated across multiple resolutions, specifically, $32\times32$, $48\times48$, $64\times64$, and $224\times224$. The first two resolutions enable us to analyze the effect of the proposed model for low resolution classification, $64\times64$ enables us to understand its performance for moderate sized images, while $224\times224$ allows us to study the effect of the proposed loss for higher resolutions as well. 

\subsection{Implementation Details}
The proposed attribute-assisted loss is implemented in PyTorch \cite{pytorch}. All pre-trained models have been taken from the Torchvision library and have been trained and tested using a Nvidia 1080Ti GPU. Each model is trained with the Stochastic Gradient Descent optimizer \cite{sgd}, with an initial learning rate of 0.01 and momentum of 0.9 for 100 epochs. The learning rate is reduced by a factor of 10 after every 20 iterations. A weight decay of 0.01 is also used as a regularizer during the training process. The weight of the attribute-assisted term ($\lambda$) is set to 0.01 after performing empirical evaluation. For smaller resolutions ($32\times32$ and $48\times48$), a batch size of 400 samples is used for training, while for larger resolutions of $64\times64$ and $224\times224$, a batch size of 100 and 50 is used, respectively. For each experiment, all images have been synthetically down-sampled to the required resolution. 

\begin{table}
\centering
\caption{Rank-1 (top-1) classification accuracy obtained on the AwA-2 dataset for different resolutions and models. \textit{Euc.} corresponds to the accuracy obtained upon matching the features obtained directly from the pre-trained model with Euclidean distance. \textit{CE Only} refers to fine-tuning the pre-trained model with the traditional Cross Entropy loss, and \textit{Proposed} refers to the accuracy obtained with the proposed attribute-assisted loss.}
\label{awa2:res}
\begin{tabular}{|c|l|c|c|c|}
\hline
\textbf{Resolution} & \textbf{Model} & \textbf{Euc.} & \textbf{CE Only} & \textbf{Proposed} \\
\hline
\hline
\multirow{3}{*}{$32\times32$} & ResNet-18 & 18.89 & 36.31 & 44.54 \\
\cline{2-5}
& ResNet-50 & 20.70	& 38.09	& \textbf{53.24}\\
\cline{2-5}
& DenseNet & 18.98	& 41.36	& 50.56 \\
\hline
\hline
\multirow{3}{*}{$48\times48$} & ResNet-18 & 27.34 & 51.60 & 61.35\\
\cline{2-5}
& ResNet-50 & 33.08 & 60.63 	&  \textbf{67.10} \\
\cline{2-5}
& DenseNet & 28.09 & 61.59 & 64.29 \\
\hline
\hline
\multirow{3}{*}{$64\times64$} & ResNet-18 & 40.03 & 62.44 & 67.47 \\
\cline{2-5}
& ResNet-50 & 48.21	& 72.90	& 75.84 \\
\cline{2-5}
& DenseNet & 44.55 &	75.94 &	\textbf{76.87} \\
\hline
\hline
\multirow{3}{*}{$224\times224$} & ResNet-18 & 86.12	& 90.29	& 91.95\\
\cline{2-5}
& ResNet-50 & 90.14	& 93.40 & 93.68 \\
\cline{2-5}
& DenseNet & 85.51 & 93.57 &	\textbf{93.73} \\
\hline
\end{tabular}
\vspace{-5pt}
\end{table}

\section{Results and Analysis}
The proposed attribute-assisted loss has been evaluated with three models and four resolutions. Results have been reported on the respective test set of the two datasets. Table \ref{awa2:res} and Table \ref{deepFashion:res} present the classification accuracies\footnote{Consistent with literature, rank-1 or top-1 accuracy has been reported, i.e. the percentage of samples correctly classified at the first rank only.} obtained by the proposed attribute-assisted loss on the AwA-2 and DeepFashion datasets, respectively. For the proposed loss, \textit{white} and \textit{clothing type (upper, lower, or full body)} attributes have been used for the two datasets, respectively. The tables also present the accuracies obtained upon performing classification with the features extracted via the pre-trained model only with Euclidean distance (Euc.), and the accuracies obtained upon fine-tuning the pre-trained model with only the traditional Cross Entropy loss (CE Only). It can be observed that the proposed loss consistently improves the classification performance for the three models, across resolutions. 

\begin{table}
\centering
\caption{Rank-1 (top-1) classification accuracy on the DeepFashion dataset for different resolutions and models. \textit{Euc.} refers to the accuracy on matching the features obtained directly from the pre-trained model with Euclidean distance. \textit{CE Only} refers to fine-tuning the pre-trained model with the traditional Cross Entropy loss, and \textit{Proposed} refers to the accuracy obtained with the proposed attribute-assisted loss.}
\label{deepFashion:res}
\begin{tabular}{|c|l|c|c|c|}
\hline
\textbf{Resolution} & \textbf{Model} & \textbf{Euc.} & \textbf{CE Only} & \textbf{Proposed} \\
\hline
\hline
\multirow{3}{*}{$32\times32$} & ResNet-18 & 34.96 & 58.64 & 63.97 \\
\cline{2-5}
& ResNet-50 & 35.56	& 59.08 & \textbf{64.02}\\
\cline{2-5}
& DenseNet &  36.36	& 60.98	& 62.49 \\
\hline
\hline
\multirow{3}{*}{$48\times48$} & ResNet-18 &  39.79 & 64.22 & 67.34\\
\cline{2-5}
& ResNet-50 & 38.04  & 66.28 	& \textbf{67.96}  \\
\cline{2-5}
& DenseNet &  40.02  & 64.69 & 67.41 \\
\hline
\hline
\multirow{3}{*}{$64\times64$} & Resnet-18 & 42.86 & 65.42 & 68.76 \\
\cline{2-5}
& ResNet-50 & 42.07 	& 67.29	& 68.74 \\
\cline{2-5}
& DenseNet & 44.02 & 67.90 & \textbf{70.49} \\
\hline
\hline
\multirow{3}{*}{$224\times224$} & Resnet-18 & 53.66 & 71.25 & 72.51 \\
\cline{2-5}
& ResNet-50 & 56.33 & 72.50	& 74.71 \\
\cline{2-5}
& DenseNet &  56.62 & 72.46  & \textbf{74.94}	  \\
\hline
\end{tabular}
\vspace{-5pt}
\end{table}

\subsection{Effect of Resolution}
Table \ref{awa2:res} and Table \ref{deepFashion:res} can be analyzed for resolution specific performance of the proposed attribute-assisted loss. On the AwA-2 dataset, a classification accuracy of 53.24\%, 67.10\%, 76.87\%, and 93.73\% is obtained for $32\times32$, $48\times48$, $64\times64$, and $224\times224$ resolutions, respectively. On the other hand, for the DeepFashion dataset, an accuracy of 64.02\%, 67.96\%, 70.49\%, and 74.94\% is obtained for the four resolutions, respectively. For smaller resolutions of $32\times32$ and $48\times48$, the proposed attribute-assisted loss consistently performs best with the ResNet-50 model, while on higher resolutions, the DenseNet-121 model performs best. 

\begin{figure} [t]
\centering
\subfloat[][Animals with Attributes 2]{\includegraphics[width = 3in]{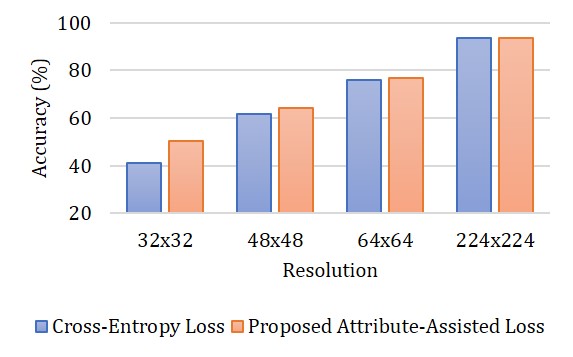}} \\
\subfloat[][Deep Fashion]{\includegraphics[width = 3in]{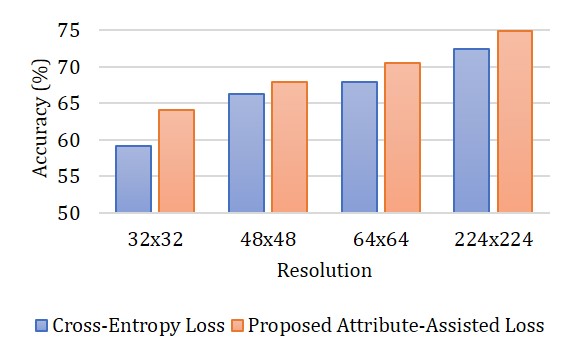}} 
\caption{Comparison of classification accuracies of the best performing model for the two datasets. Accuracy obtained via the Cross Entropy loss is compared with the proposed attribute-assisted loss. It is interesting to note that the effect of the attribute information is more prominent in low resolution images, as compared to the higher ones.}
\label{fig:improvement}
\vspace{-5pt}
\end{figure}

The two tables can also be read from right to left in order to understand the \textit{ablation study} on the proposed loss. The proposed technique utilizes both, the classification loss and the attribute loss for learning the network. On the other hand, `CE only' uses only the classification loss. Therefore, the improvement observed across the two columns is attributed to the additional attribute assistance provided during learning. %It is interesting to observe that the contribution of the attribute-assisted loss is more prominent in lower resolutions as compared to higher resolutions. 
Fig. \ref{fig:improvement} presents the improvement in accuracy observed for the best performing models for both the datasets, across the resolutions. For example, for the AwA-2 dataset, an improvement of almost 15\% is observed at a smaller resolution of $32\times32$ with the ResNet-50 model, whereas an improvement of less than 1\% is observed at a higher resolution of $224\times224$. Similarly, for DeepFashion dataset, lower resolution classification benefits more from the proposed attribute-assisted loss as compared to higher resolution classification. We believe that at lower resolutions, where the information content is limited, additional information helps in learning better features, which in turn improves the classification performance. On the other hand, at higher resolutions, since the information content is as it is high, the effect of the attribute-assisted loss is less pronounced. 

It is also interesting to observe that the improvement in accuracy is more prominent for the AwA-2 dataset, as compared to the DeepFashion dataset. It is important to note that the number of training samples for the AwA-2 dataset and DeepFashion differ almost by a ratio of 8 (AwA-2 has only around 26K training samples, while DeepFashion has over 200K samples). It is our assertion that for the smaller dataset of AwA-2, the proposed attribute-assisted loss also acts as a regularizer, thus preventing the model from overfitting on the training data. This also results in higher improvement at lower resolutions, where less information content coupled with relatively less data renders the problem further challenging.

\begin{table*}
\centering
\caption{Rank-1 (top-1) classification accuracy on the Animals with Attributes-2 dataset for $32\times32$ and $48\times48$ resolutions. Results are demonstrated with four different attributes - both visual and non-visual. Comparison with the traditional Cross Entropy loss (CE only) demonstrate the effectiveness of the proposed loss. }
\label{awa2:diffAttr}
\begin{tabular}{|c|l|c|c|c|c|c|}
\hline
\multirow{2}{*}{\textbf{Resolution}} & \multirow{2}{*}{\textbf{Model}} & \multirow{2}{*}{\textbf{CE Only}} & \multicolumn{4}{c|}{\textbf{Proposed Attribute-Assisted Loss}} \\
\cline{4-7} 
  &   &   & \textbf{White} & \textbf{Patches} & \textbf{Domestic} & \textbf{Forest} \\
\hline
\hline
\multirow{3}{*}{$32\times32$} & ResNet-18 & 36.31 & 44.54 & 43.94 & 44.25 & {48.63}\\
\cline{2-7}
& ResNet-50 & 38.09 & \textbf{53.24}	& \textbf{52.60}	& \textbf{51.84} & \textbf{52.80}\\
\cline{2-7}
& DenseNet-121 & 41.36& {50.56}	& 49.16	& 46.36 & 49.11 \\
\hline
\hline
\multirow{3}{*}{$48\times48$} & ResNet-18 & 51.60 & {61.35} & 60.72 & 59.63 & 57.36\\
\cline{2-7}
& ResNet-50 & 60.63 & \textbf{67.10} & 66.22 & 64.06 & 65.80\\
\cline{2-7}
& DenseNet-121 & 61.59 & 64.29 & \textbf{66.83}  & \textbf{67.01} & \textbf{66.52} \\
\hline
\end{tabular}
\end{table*}

\begin{table}
\centering
\caption{Comparison of the proposed attribute-assisted loss on DenseNet-121 with state-of-the-art techniques on the DeepFashion dataset and the AwA-2 dataset for $224\times224$ resolution. Due to the same protocol, comparison has been performed with the ranks reported in the literature.}
\label{deepFashion:sota}
\begin{tabular}{|l|c|c|}
\hline
\multicolumn{3}{|c|}{\textbf{DeepFashion Dataset}} \\
\hline
\hline
\textbf{Algorithm} & \textbf{Top-3} & \textbf{Top-5}  \\
\hline
WTBI (2012) \cite{eccv12} & 44.73 & 66.26 \\
\hline
DARN (2015) \cite{iccv15} & 59.48 & 79.58 \\
\hline
FashionNet+Joints (2016) \cite{deepFashion} & 72.30 & 81.52 \\
\hline
FashionNet+Poselets (2016) \cite{deepFashion} & 75.34 & 85.87 \\
\hline
FashionNet (2016) \cite{deepFashion} & 82.58 & 90.17 \\
\hline
Lu \textit{et al.} (VGG-16) (2017) \cite{lu2017fully} & 86.72 & 92.51 \\
\hline
Corbiere \textit{et al.} (2017) \cite{corbiere2017leveraging} & 86.30 & 92.80 \\
\hline
SEResNeXt50+OC (2019) \cite{park2019study} & 88.42 & 93.93 \\
\hline
\textbf{Proposed Attribute-Assisted Loss} & {\textbf{91.05}} & {\textbf{95.35}} \\
\hline
\hline
\multicolumn{3}{|c|}{\textbf{AwA-2 Dataset}} \\
\hline
\hline
\textbf{Algorithm} & \multicolumn{2}{c|}{\textbf{Top-1}} \\
\hline
Xiao \textit{et al.} (2020) \cite{xiao2020novel} & \multicolumn{2}{c|}{86.79} \\
\hline
\textbf{Proposed Attribute-Assisted Loss} & \multicolumn{2}{c|}{\textbf{93.73}} \\
\hline
\end{tabular}
\vspace{-10pt}
\end{table}

\subsection{Comparison with Existing Algorithms}
Since low resolution fine-grained classification has garnered limited attention in the research community, this is the first work presenting the baselines for low resolution classification on the AwA-2 and DeepFashion datasets. Traditionally, these datasets have mostly been used for reporting results for zero-shot or few-shot protocols. In the literature, research has focused on high resolution fine-grained classification on the DeepFashion dataset as well. In this experiment, to compare with existing techniques, results are obtained for the proposed attribute-assisted loss using Densenet-121 on $224\times224$ resolution using the protocol given in Section \ref{sec:protocol}. As per the protocol, Table \ref{deepFashion:sota} presents the top-3 and top-5 accuracy of the proposed model along with other comparative techniques on DeepFashion database\footnote{Consistent with literature, top-3 and top-5 accuracy has been reported, i.e. percentage of correct classifications within the top 3 or 5 ranks, respectively.}. It can be observed that the proposed loss achieves the best performance by demonstrating an improvement of over 8\% and 3\% for the top-3 and top-5 accuracies, respectively. One of the recent comparative techniques on this dataset, FashionNet, uses a CNN architecture to predict the attribute, landmark, and category of the input image. On the other hand, the proposed attribute-assisted loss pushes the model to learn features separable on a given attribute. The benefit of providing ancillary information, as opposed to performing classification on the attribute, can be observed in the improved performance of the proposed loss. Further, the pre-defined protocol on the AwA-2 dataset is for zero-shot learning only, and thus there exist limited comparative results for classification. In 2020, Xiao \textit{et al.} \cite{xiao2020novel} proposed a novel pooling block which obtains 86.79\% classification performance on the AwA-2 dataset. In comparison, the proposed attribute-assisted loss demonstrates an improvement of around 7\% by obtaining 93.73\% (Table \ref{deepFashion:sota}).

\subsection{Effect of $\lambda$}
\begin{figure}
\centering
\includegraphics[trim={0.5cm 0.2cm 0.5cm 0},clip, width=3.2in]{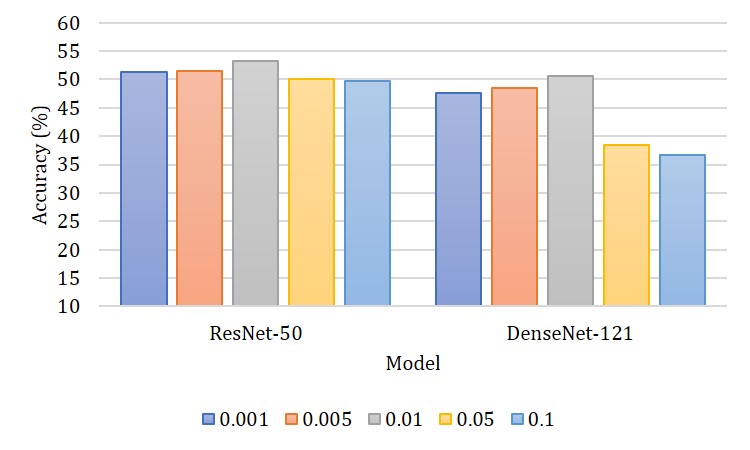}
\caption{Effect of $\lambda$, weight of the attribute-loss. Analysis is performed on $32\times32$ resolution, with ResNet-50 and DenseNet-121, for the AwA-2 dataset.}
\label{fig:weight}
\end{figure}

As mentioned previously, the proposed attribute-assisted loss contains two main components: (i) the classification loss and (ii) the attribute loss. The classification loss corresponds to the traditional Cross Entropy loss used for classifying the input data into one of the classes. The attribute loss works with the ancillary information and attempts to minimize mis-classifications between classes of different attributes. As shown in Eq. \ref{eq:prop}, $\lambda$ controls the contribution of the attribute-assisted loss in the final loss function. In this subsection, the contribution of the attribute-assisted loss is analyzed based on the weight parameter. Experiments have been performed on the AwA-2 dataset, with low resolution data ($32\times32$). Fig. \ref{fig:weight} demonstrates the effect of varying the weight parameter ($\lambda$) on the classification accuracy. Experiments have been performed on the top two models: Resnet-50 and Densenet-121. A similar trend is observed for both the models, where increasing the weight value results in an increase of the classification performance, till a particular value. At a weight of 0.01, both the models reach the peak performance, beyond which the performance of the model declines. These trends are intuitive in nature, for instance, at a smaller weight value, the contribution of the attribute-assisted term is less, whereas on larger weight values, the contribution of the attribute loss offsets the classification loss, thereby resulting in a drop in the performance.  

%\begin{table}
%\caption{Animals with Attributes-2 - results with different attributes.}
%\label{awa2:diffAttr}
%\begin{tabular}{|c|l|c|c|c|c|}
%\hline
%\textbf{Resltn.} & \textbf{Model} & \textbf{White} & \textbf{Patches} & \textbf{Domestic} & \textbf{Forest} \\
%\hline
%\hline
%\multirow{3}{*}{$32\times32$} & Resnet-18 & 44.54 & 43.94 & 44.25 & \textbf{48.63}\\
%\cline{2-6}
%& Resnet-50 & \textbf{53.24}	& 52.60	& 51.84 & 52.80\\
%\cline{2-6}
%& Densenet & \textbf{50.56}	& 49.16	& 46.36 & 49.11 \\
%\hline
%\hline
%\multirow{3}{*}{$48\times48$} & Resnet-18 & \textbf{61.35} & 60.72 & 59.63 & 57.36\\
%\cline{2-6}
%& Resnet-50 & \textbf{67.10} & 66.22 & 64.06 & 65.80\\
%\cline{2-6}
%& Densenet & 64.29 & {66.83}  & \textbf{67.01} & 66.52 \\
%\hline
%\end{tabular}
%\end{table}

\subsection{Evaluation with Different Attributes} 
The proposed attribute-assisted loss has also been evaluated with different attributes. Since the DeepFashion dataset contains only a single set of class-wise attributes, experiments are performed on the AwA-2 dataset only. Four different attributes of \textit{white}, \textit{patches}, \textit{domestic}, and \textit{forest} have been considered for evaluation, with lower resolutions of $32\times32$ and $48\times48$ resolutions. Table \ref{awa2:diffAttr} presents the performance of the proposed attribute-assisted loss with different attributes, along with the performance of the traditional Cross Entropy loss (CE only). Table \ref{awa2:diffAttr} shows that the proposed attribute-assisted loss is able to learn improved classifiers with different types of attributes. For example, while $white$ and $patches$ provide explicit visual information about the animal, $domestic$ and $forest$ can be viewed as non-visual cues. The proposed attribute-assisted loss is able to utilize both these categories of information for enhancing the performance.

\section{Conclusion}

Fine-grained classification has garnered substantial attention over the past few years, with majority of the algorithms focusing on high resolution images only. This research proposes a novel attribute-assisted loss for the challenging problem of low resolution fine-grained classification. The proposed loss utilizes attributes as additional information, thereby facilitating a given model to learn attribute-level discriminative features while performing classification. One of the key highlights of the proposed loss is its architecture agnostic behavior, and the ability to incorporate non-visual attributes as well. The efficacy of the proposed attribute-assisted loss is demonstrated with three networks (ResNet-18, ResNet-50, and DenseNet-121) on two publicly available datasets (DeepFashion and AwA-2 datasets), for four varying resolutions from $32\times32$ to $224\times224$. The improved performance obtained across different resolutions strengthens its usage for low resolution fine-grained classification. As future work, the proposed technique can be extended to handle multiple attributes during learning.

\section{Acknowledgement}

S. Nagpal is supported via the TCS PhD fellowship and M. Vatsa is partially supported through the Swarnajayanti Fellowship by the Government of India.

\bibliographystyle{IEEEtran}
\bibliography{ijcnn_v2}

\end{document}